\documentclass{article}

\usepackage[preprint]{neurips_2019}

 \usepackage{natbib}
\setcitestyle{authoryear,round, citesep={;},aysep={,},yysep={;},sort&compress, longnamesfirst}
 
\usepackage[utf8]{inputenc} 
\usepackage[T1]{fontenc}    
\usepackage{hyperref}       
\usepackage{url}            
\usepackage{booktabs}       
\usepackage{color}
\usepackage{amsfonts}
\usepackage{amsmath} 
\usepackage{amssymb}
\usepackage{nicefrac}       
\usepackage{microtype}      
\usepackage{lipsum}
\usepackage{graphicx}
\usepackage{caption}
\usepackage{subcaption}
\usepackage[linesnumbered,ruled,vlined]{algorithm2e}

\title{Nonlocal optimization of binary neural networks}
\newcommand{\I}{\mathbb{I}}

\author{%
  Amir Khoshaman \thanks{Authors contributed equally to this work.} \\
  Borealis AI\\
  \And
  Giuseppe Castiglione\footnotemark[1] \\
  Borealis AI\\
   \And
   Christopher Srinivasa \\
   Borealis AI \\
}

\begin{document}

\maketitle

\begin{abstract}

We explore training Binary Neural Networks (BNNs) as a discrete variable inference problem over a factor graph. We study the behaviour of this conversion in an under-parameterized BNN setting and propose stochastic versions of Belief Propagation (BP) and Survey Propagation (SP) message passing algorithms to overcome the intractability of their current formulation. Compared to traditional gradient methods for BNNs, our results indicate that both stochastic BP and SP find better configurations of the parameters in the BNN.

\end{abstract}

\section{Introduction}
Belief propagation (BP) is a message passing algorithm used to perform efficient inference and marginalization on probabilistic graphical models over a large set of interacting variables that have a tree-structure graph. BP works efficiently by utilizing the distributive law~\citep{koller2009probabilistic} to reduce the exponential complexity of inference or marginalization. Loopy BP (LBP) is an iterative message updating procedure that can be applied to loopy graphs. LBP can converge to one of the many possible fixed points in a loopy graph~\citep{yedidia2005constructing}. These fixed points are often satisfactory approximations to the true values of interest~\citep{murphy_graph}.
Survey propagation is another message passing algorithm extending LBP that aims to capture all the fixed points of the BP algorithm weighed by their corresponding Bethe partition function~\citep{mezard2009information}. 

The message passing algorithms can be cast conveniently on factor graphs~\citep{kschischang2001factor}, which are bipartite graphs comprising of variables of interest and factors that capture the local relations among the variables. Due to intractability of high-degree factors, message passing algorithms
have only been generally studied in cases where each factor is connected to a small subset of variables, with exceptions involving certain classes of factors~\citep{SrinivasaMMP, braunstein2006learning, pmlr-v9-tarlow10a}. For instance, SP has mostly been applied in constraint satisfaction problems (CSPs)~\citep{braunstein2005survey} and recently in Ising problems~\citep{pmlr-v51-srinivasa16}. Extending BP and SP to high degree factors can open up avenues to study some interesting problems such as BNNs from a different perspective. This provides several advantages with respect to gradient-based approaches. First, one can use an inherently nonlocal optimization method such as SP to obtain solutions that are not accessible to local methods. Moreover, these methods naturally  accommodate inherently discrete loss functions.  These approaches lend themselves naturally to a Bayesian setting, where distributions over parameters, rather than point estimates, are obtained.
BNNs are of much interest since they  address some of the vast memory and computing power demands of deep neural networks. They are suited to real-time applications where neural networks are embedded in an energy-constrained environment~\citep{peters2018probabilistic}. Recent works on BNNs are gradient-based and use stochastic gradient descent (SGD) for training. Binarizing the weights can happen either after training~\citep{Louizos2017BayesianCF}, or during training~\citep{Ullrich2017, peters2018probabilistic, Achterhold2018VariationalNQ}.

The contributions of this work are as follows: In Sec.~\ref{sec:Stoch_BP}, we provide a probabilistic interpretation of BP and propose a stochastic extension that allows (arbitrary) factors with any degree of connectivity. In Sec.~\ref{sec:Stoch_SP}, we propose two extensions of SP with stochastic messages that extend SP to high degree factors and variables. We demonstrate in Sec.~\ref{sec:exp:consistency} that our proposed models are consistent with exact BP and SP in small BNNs. In Sec.~\ref{sec:exp:toy_datasets}, we study the performance of our models in under- and over-parameterized conditions and demonstrate that they are able to obtain better solutions than gradient descent (GD). Finally in Sec.~\ref{sec:exp:mnist}, we examine the performance of our models in high degree factors including multi layered perceptrons (MLPs) and convolutional networks on MNIST dataset and show that they perform well compared with SGD under these conditions.

\section{Background}
Here, we define the general problem setting, review BP and SP and set the notation that is used throughout the following sections.
Consider a function $F(\underline{w})$ over $\underline{w} = \{w_1, \dots, w_N\}$ with $|w_i|=K$, \emph{i.e.}, each $w_i$ has cardinality $D$ (assuming one of $D$ possible values). We adopt the convenient notation used in~\citet{pmlr-v51-srinivasa16}, where underlined variables represent sets. Marginalization of $F(\underline{w})$ over a set of variables indexed by $\alpha \subset\{1, \dots, N\} $, $\underline{w}_\alpha$, involves a sum with $D^{N-|\alpha|}$ terms ($|\alpha|$ is the cardinality of $\underline{w}_\alpha$) and is intractable.
\subsection{Belief Propagation}
\label{sec:BP}
When $F(\underline{w})$ factorizes as a product of local ``factors'' $f_I(\underline{w}_I)$, 
the distributive law~\citep{bishop2006pattern}, $ab + ac = a(b+c)$ can be used to solve the marginalization problem. The problem can be represented as messages passed on a factor graph~\citep{kschischang2001factor} which is a bipartite graph comprising of factors, $\underline{f}_I$, and variables, $\underline{w}_i$, indexed by upper and lower letters, respectively. The computational gain is achieved by first applying a summation to the factors (instead of the whole $F(\underline{w})$) and then accumulating the results as products on variables. This ``message-passing'' algorithm leads to two types of messages, the variable to factor and factor to variable messages. The message, ${m}_{I\rightarrow i}(w_i)$, 
passed from factor $I$ to variable $i$ is defined as:
\begin{equation}
    {m}_{I\rightarrow i}(w_i; m_{\partial I\setminus i \rightarrow I}) \triangleq \sum_{\underline{w}_{\setminus i}}f_I(\underline{w}_I) \prod _{j \in \partial{I} \setminus{i}} m_{j\rightarrow I}(w_j), \label{eq:f2var_orig}
\end{equation}
where the symbol $\partial [.]$ denotes the set of nodes (variables or factors) adjacent to node $[.]$, $\partial I\setminus i \rightarrow I$ represents all variables adjacent to factor $I$ except $i$. The message from variable $i$ to factor $I$ is defined as:
\begin{equation}
    {m}_{i\rightarrow I}(w_i; m_{\partial i\setminus I \rightarrow i}) \triangleq \prod _{J \in \partial{i} \setminus{I}} m_{J\rightarrow i}(w_j).
    \label{eq:var2f_orig}
\end{equation}

In order to make the equations less cluttered, the functional dependence of the messages on $m_{\partial i\setminus I \rightarrow i}$ and $m_{\partial i\setminus i \rightarrow I}$ are suppressed when no confusion arises.
The belief of variable $i$ is defined as $  {m}_{i}(w_i; m_{\partial i \rightarrow i}) \triangleq \prod _{J \in \partial{i}} m_{J\rightarrow i}(w_j)$ which is equal to the true marginal in tree-structured factor graphs.
BP works by choosing a variable as the root node in the graph, and subsequently passing messages between the variables and factors using Eqs.~\ref{eq:f2var_orig}-\ref{eq:var2f_orig}. When the factor graph is a tree, the computed marginal is exact. LBP can be used to perform approximate inference on loopy graphs by initializing the messages and iteratively applying Eqs.~\ref{eq:f2var_orig}-\ref{eq:var2f_orig}. 

\subsection{Survey Propagation}
\label{sec:SP}
It is demonstrated by~\citet{Heskes2002StableFP} that when LBP converges (a stable fixed point) on a loopy factor graph, a local minimum of the Bethe approximation to the partition function is obtained. A factor graph with loops can have many BP fixed points. SP is a powerful extension of BP designed to keep track of an exponential number of fixed points in LBP~\citep{Ravanbakhsh2014RevisitingAA}. This is achieved by assuming that the joint distribution represented by the factor graph can be approximated by a sum over the joint distributions obtained by fixed points of LBP. Therefore, marginalization in SP involves a sum over all corresponding variables for a given fixed point, and eventually, summing over all the fixed points.
The conditions for validity of SP approximation correspond to a phase in replica symmetry breaking~\citep{mezard2009information} known as the clustering phase which is not easy to check.  \citet{pmlr-v51-srinivasa16} demonstrated that regardless of the validity of assumptions, SP can operate on a wide range of problems and obtain solutions that are not accessible to BP. Similar to BP, SP comprises of two sets of messages.
The variable to factor message is defined as~\citep{pmlr-v51-srinivasa16}:
\begin{align}
   {M}_{i \rightarrow I}({m}_{i \rightarrow I}=m_0) & \triangleq
    \int_{{\underline{m}'}_{\partial i \setminus I \rightarrow i}} \Bigg[
   \mathbb{I}\left[ m_0 = 
   {m}_{i\rightarrow I}(.; m'_{\partial i\setminus I \rightarrow i})
   \right] \times \nonumber \\
  &{m}_{i\rightarrow I}(.; m'_{\partial i\setminus I \rightarrow i})(\emptyset)
   \prod_{J \in \partial i\setminus I}{M}_{J \rightarrow I}({m'}_{J \rightarrow i})
   \Bigg].
   \label{eq:SP_var2f_orig}
\end{align}
Here, the integration is over the space of all possible BP factor ($\partial i \setminus I$) to variable ($i$) message functions $\underline{m}'_{\partial i \setminus I \rightarrow i}$. ${M}_{i \rightarrow I}({m}_{i \rightarrow I}=m_0)$ can be interpreted as the message that variable $i$ sends to factor $I$ signaling that the corresponding variable to factor BP message is equal to $m_0$. The indicator function $\I [.]$ constrains the integration region to domains that respect the corresponding BP update. Note that ${m}_{i\rightarrow I}(.; m'_{\partial i\setminus I \rightarrow i})(\emptyset)\triangleq \sum _{w_i}{m}_{i\rightarrow I}(w_i; m'_{\partial i\setminus I \rightarrow i})$. Similarly, the factor to variable message is defined as (with similar notation and definitions):
\begin{align}
   {M}_{I \rightarrow i}({m}_{I \rightarrow i}=m_0) & \triangleq
    \int_{{\underline{m}'}_{\partial I \setminus i \rightarrow I}} \Bigg[
   \mathbb{I}\left[ m_0 = 
   {m}_{I\rightarrow i}(.; m'_{\partial I\setminus i \rightarrow I})
   \right] \times \nonumber \\
  &{m}_{I\rightarrow i}(.; m'_{\partial I\setminus i \rightarrow I})(\emptyset)
   \prod_{j \in \partial I\setminus i}{M}_{j \rightarrow I}({m'}_{j \rightarrow I})
   \Bigg].
   \label{eq:SP_f2var_orig}
\end{align}
In SP, these two types of messages are updated iteratively until convergence and eventually used to calculate the marginals (See Sec.~\ref{sec:Stoch_SP}).

\section{Proposed approach}
\label{sec:Method}

\subsection{Stochastic BP (SBP)}
\label{sec:Stoch_BP}
The factor to variable message update in BP, ${m}_{I\rightarrow i}(w_i)$, (\emph{i.e.}, Eq.~\ref{eq:f2var_orig}) involves a sum over $\underline{w}_{\setminus i}$. However, one only needs to consider the variables $\underline{w}_I$, since $f_I(\underline{w}_I)$ does not depend on the other variables $\underline{w}_{\setminus i \setminus I}$. Nonetheless, this sum involves $D^{|\underline{w}_I|}$ terms, which is only tractable for low-degree factors. In this section, we propose a stochastic formulation of BP that allows 
high degree factors.

Notice that the variable to factor messages that appear in Eq.~\ref{eq:f2var_orig} are all positive numbers, therefore, we can convert $\prod _{j \in \partial{I} \setminus{i}} m_{j\rightarrow I}(w_j)$ to a joint probability distribution $p({w}_{\partial{I} \setminus{i}}) \triangleq \prod _{j \in \partial{I} \setminus{i}} {p}_{j\rightarrow I}(w_j)$ where ${p}_{j\rightarrow I}(w_j) \triangleq \frac{{m}_{j\rightarrow I}(w_j)}{{m}_{j\rightarrow I}(\emptyset)}$ and ${m}_{j\rightarrow I}(\emptyset) \triangleq \sum_{w_j}{m}_{j\rightarrow I}(w_j)$. This transformation is justified since we are only interested in the ratios of ${m}_{I\rightarrow i}(w_i)$ for the possible $D$ values of $w_i$ and this transformation multiplies all the possible values of the message by the same quantity.

Therefore, Eq.~\ref{eq:f2var_orig} can be rewritten as:
\begin{equation}
    \tilde{p}_{I\rightarrow i}(w_i;p_{\partial I\setminus i \rightarrow I}) = \sum_{\underline{w}_{\setminus i}}f_I(\underline{w}_I) \prod _{j \in \partial{I} \setminus{i}} {p}_{j\rightarrow I}(w_j)
    , \label{eq:f2var_norm}
\end{equation}
where $\tilde{p}_{I\rightarrow i}(w_i) \propto {m}_{I\rightarrow i}(w_i)$.

 Our proposed approach is to regard Eq.~\ref{eq:f2var_norm} as expectation value of $f_I(\underline{w}_I)$ with respect to the joint probability distribution $p({w}_{\partial{I} \setminus{i}})$.
The value of the message can be estimated using a Monte Carlo (MC) approach, \emph{i.e.}, drawing samples from the joint distribution:
\begin{equation}
        \tilde{p}_{I\rightarrow i}(w_i;p_{\partial I\setminus i \rightarrow I}) = \mathbb{E}_{p({w}_{\partial{I} \setminus{i}})} \left[f_I(\underline{w}_I)  \right] \approx \frac{1}{L} \sum_{m=1}^{L}f_I({\underline{w}_{I, m}}),
    \label{eq:f2var_stoch}
\end{equation}
where $m$ indexes over the $L$ draws from the joint distribution.
We call this model stochastic BP (SBP). It is not related to \citep{Noorshams_2013}, where a stochastic approach is employed to deal with high values of $D$. 
Algorithm~\ref{alg:bp_f2var} illustrates this approach when the variables are binary ($w_i \in \{ 0, 1\}$). This corresponds to a joint distribution over a factorial Bernoulli distribution.
The algorithm loops over all variables $i$ adjacent to a factor $I$ (lines 1-11). It also loops over $L$ MC samples drawn from the joint probability distribution defined by the variable to factor messages (lines 2-7). The $i$th elements, $w_i$ of the drawn samples are set to $1$ and $0$ at lines 4 and 6, respectively. These values are used to obtain MC estimates of the unnormalized probabilities of sending a $1$ or a $0$ from $I$ to $i$ in lines 8-9. The normalized probability of sending a $1$ is calculated in line 10.
\begin{algorithm}[!htbp]
  \SetAlgoLined
\KwIn{Normalized messages $\{{p}_{i \rightarrow I}(w_i)\}_{\{i \in \partial I \}}$ incident on factor $I$}
\KwOut{$\{{p}_{I \rightarrow i}(w_i=1) \}_{\{i \in \partial I \}}$, $\tilde{p}_{I \rightarrow i}(w_i=1)$ and $\tilde{p}_{I \rightarrow i}(w_i=0)$ from factor $I$ }
 \For{$i \in \partial I$}{
 \For{$m \in \{1, \dots, L\}$}{
 $\underline{w}^{1}_{I, m} \sim \{\prod _{j \in \partial{I} } {p}_{j\rightarrow I}(w_j)  \} \quad $
 
 ${w}^{1}_{i, m} \leftarrow 1 \quad$
 
  $\underline{w}^{0}_{I, m} \sim \{\prod _{j \in \partial{I} } {p}_{j\rightarrow I}(w_j)  \} \quad $
 
 ${w}^{0}_{i, m} \leftarrow 0 \quad$
 }
  
$\tilde{p}_{I \rightarrow i}(w_i=1) \leftarrow \frac{1}{L} \sum_{m=1}^{l}f_I({\underline{w}^{1}_{I, m}})$
  
$\tilde{p}_{I \rightarrow i}(w_i=0) \leftarrow \frac{1}{L} \sum_{m=1}^{l}f_I({\underline{w}^{0}_{I, m}})$
  
${p}_{I \rightarrow i}(w_i=1) \leftarrow \frac{\tilde{p}_{I \rightarrow i}(w_i=1)}{\tilde{p}_{I \rightarrow i}(w_i=0)+\tilde{p}_{I \rightarrow i}(w_i=1)}$
 }
 \caption{Stochastic BP factor to variable update subroutine for binary variables: \texttt{StochasticBPF2Var}$(\{{p}_{i \rightarrow I}(w_i)\}_{\{i \in \partial I \}})$.}
 \label{alg:bp_f2var}
\end{algorithm}
\subsection{Stochastic SP (S$^3$P and S$^4$P)}
\label{sec:Stoch_SP}
The integrals in the SP message updates are converted into  sums by keeping histograms of BP messages. With $K$ bins, the sums involve $K^{|\partial I|-1}$ and $K^{|\partial i|-1}$ terms, for variable to factor and factor to variable messages, respectively. These sums are both intractable for high-degree nodes. The exponential complexity of the variable to factor messages can be reduced to $\mathcal{O}(K^2|\partial i|)$ by performing FFT, multiplying the signals, and then doing an inverse FFT, see \citep{pmlr-v51-srinivasa16} for details. This process involves padding the original messages from size $K$ to $\approx |\partial i|K$ in order to avoid aliasing. Since $|\partial i|$ represents the number of examples in a dataset (see Sec.~\ref{sec:exp}), it is not practical for large datasets. Here, we provide probabilistic interpretations of both these messages to obtain stochastic estimates of the messages in linear time.

Consider the SP factor to variable update, Eq.~\ref{eq:SP_f2var_orig}. By defining $\tilde{S}_{j \rightarrow I}({p}_{j \rightarrow I}) \propto{{M}_{j \rightarrow I}({p}_{j \rightarrow I})}$ with
 ${S}_{j \rightarrow I}({p}_{j \rightarrow I}) \triangleq \frac{\tilde{S}_{j \rightarrow I}({p}_{j \rightarrow I})}{\sum_{{p}_{j \rightarrow I}} \tilde{S}_{j \rightarrow I}({p}_{j \rightarrow I})}$
 , we can regard the factor to variable update as an expectation under the joint distribution $S({p}_{\partial I\setminus i}) \triangleq \prod_{j \in \partial I\setminus i}{S}_{j \rightarrow I}({p}_{j \rightarrow I})$. Note that each ${S}_{j \rightarrow I}({p}_{j \rightarrow I})$ is a categorical distribution (with cardinality $K$). The resulting factor to variable message is:
\begin{equation}
    \tilde{S}_{I \rightarrow i}({p}_{I \rightarrow i}= p_k) =
    \mathbb{E}_{S({p}_{\partial I\setminus i})}\Bigg[
       \mathbb{I}\left[ p_k = {p}_{I\rightarrow i}(.; p_{\partial I\setminus i \rightarrow I})\right]
   {p}_{I\rightarrow i}(.; p_{\partial I\setminus i \rightarrow I})
   (\emptyset)
    \Bigg],
    \label{eq:SP_f2var_stoch}
\end{equation}
where ${p}_{k} \in [0, 1]$ is one of the possible $K$ values that ${p}_{I \rightarrow i}$ can take. ${p}_{I\rightarrow i}(.; p_{\partial I\setminus i \rightarrow I})
   (\emptyset)$ is estimated stochastically using Algorithm~\ref{alg:bp_f2var}. Algorithm~\ref{alg:sp_f2var} shows how this approach works. The algorithm receives variable to factor surveys (SP messages) as input $\{{S}_{i \rightarrow I}({p}_{{i \rightarrow I}})  \}_{\{i \in \partial I\}}$.
 The MC estimate is done over $L_{sp}$ samples (lines 1-9). A  variable to factor probability distribution is sampled from the surveys incident on factor $I$. This probability distribution is fed into the \texttt{StochasticBPF2Var} subroutine to obtain the normalized and unnomarlized probabilities (line 3). These unnormalized probabilities are used to calculate the local partition function for each variable (line 5), whereas the normalized probability is used to find the bin $k$ that corresponds to the given probability (line 6), where $\texttt{bin}(p_k) \triangleq k$. The message corresponding to this bin is subsequently updated (line 7). Finally, the normalized messages are calculated (lines 10-14).
   A similar approach can be used to update the variable to factor messages (See Appendix~\ref{sec:app:var2f_sp}). We call our model S$^4$P (the first 3 S's correspond to the three sources of stochasticity from the factor to variable, variable to factor and BP messages, while the last S stands for survey) as opposed to S$^3$P that uses the FFT trick to calculate the variable to factor messages.
   
   Since SP is a generalization of the BP algorithm, we call a combination of SBP and S$^4$P the Stochastic Nonlocal Message Passing (SNMP) model. SNMP starts by running the faster BP first, and if all the constraints are not satisfied, S$^4$P is called.

\begin{algorithm}[!htbp]
  \SetAlgoLined
\KwIn{Normalized messages $\{{S}_{i \rightarrow I}({p}_{{i \rightarrow I}})  \}_{\{i \in \partial I\}}$ incident on factor $I$}
\KwOut{Normalized outgoing messages $\{{S}_{I \rightarrow i}({p}_{{i \rightarrow I}})  \}_{\{i \in \partial I\}}$ from factor $I$ }
\For{$m \in \{1, \dots L_{\text{sp}}\}$}
{
    ${ p}_{ \partial I \rightarrow I} \sim 
    \prod_{j \in \partial I}{S}_{j \rightarrow I}({p}_{j \rightarrow I})
    $

    ${ p}_{I \rightarrow i}(w_{\partial I}=1),  {\tilde p}_{I \rightarrow i}(w_{\partial I}=1),  {\tilde p}_{I \rightarrow i}(w_{\partial I}=0)$ $\leftarrow$
    \texttt{StochasticBPF2Var}$({ p}_{ \partial I \rightarrow I})$
    
        \For{$i \in \partial I$}{
        
        ${p}_{I \rightarrow i}(\emptyset) \leftarrow {\tilde p}_{I \rightarrow i}(w_i=0)+{\tilde p}_{I \rightarrow i}(w_i=1)$
        
        $
        {k} \leftarrow \texttt{bin}({ p}_{I \rightarrow i}(w_i=1))
        $
        
        $
        \tilde{S}_{I \rightarrow i}(p_k) \leftarrow \tilde{S}_{I \rightarrow i}(p_k) + {p}_{I \rightarrow i}(\emptyset)
        $

 }
 }
 \For{$i\in \partial I$}{
 \For{$k \in \{1, ..., K\}$}{
 ${S}_{I \rightarrow i}({p}_{k}) \leftarrow \frac{\tilde{S}_{I \rightarrow i}({p}_{k})}{\sum_{k=1}^{K} \tilde{S}_{I \rightarrow i}({p}_{k})}$
 }
 }
 \caption{SP Stochastic factor to variable subroutine. 
 }
 \label{alg:sp_f2var}
\end{algorithm}
\subsection{Problem set-up}
\label{sec:method:problem_setup}
Consider a training set of binary vectors $\underline{x}$ and labels, $\underline{y}$, comprising of $M$ examples. We assume binary labels, $y_I \in \{0, 1\}$, for simplicity in this work, though any number of labels can also be studied using this approach. Assume a binary neural network with weights $\underline{w}$ that predicts label $g_{\underline{w}}(x_I)$ given input $x_I$. $g_{\underline{w}}(.)$ can be any function including MLPs with any type of activation. We assume that the last layer of MLP is a $\texttt{sign}$ function. The training of the MLP can be posed as a CSP problem using a (high degree) factor graph containing $M$ factors and $N \triangleq |\underline{w}|$ variables:
$F(\underline{w})=\prod_{I=1}^{M}f_I(\underline{w})$ and $f_I(\underline{w})\triangleq \mathbb{I}\left[g_{\underline{w}}(x_I)=y_I \right]$. We have considered the general case where each factor $f_I$ depends on all the variables, \emph{i.e.}, $\underline{w}_I=\underline{w}$. With this set-up, BP and SP algorithms can be employed to calculate the marginals of the weights of the neural networks that satisfy all the instances on the training set. We relax the strict requirement of satisfying all the training examples by adopting a Max-Sat strategy~\citep{Chieu2009RelaxedSP}: in sum-product forms of BP and SP, $F(\underline{w})$ is zero even when one of the factors is not satisfied. This makes the algorithm attempt to satisfy all the instances and not directly discriminate between cases that cannot satisfy one or more of the constraints. We relax this by writing $F(\underline{w}) \propto \exp(-E(\underline{w}))$, where the energy is defined as $E \triangleq -\sum_I E_I(\underline{w}))$ with $E_I(\underline{w})\triangleq -\beta(1-f_I(\underline{w}))$. Here, $\beta$ is the inverse temperature for a Boltzmann distributions in the language of statistical mechanics (assuming Boltzmann constant is unity). A high value of $\beta$ penalizes the unsatisfied factors more severely.
\section{Experiments}
\label{sec:exp}
\begin{figure}[!htbp]
    \centering
    \includegraphics[width=\textwidth]{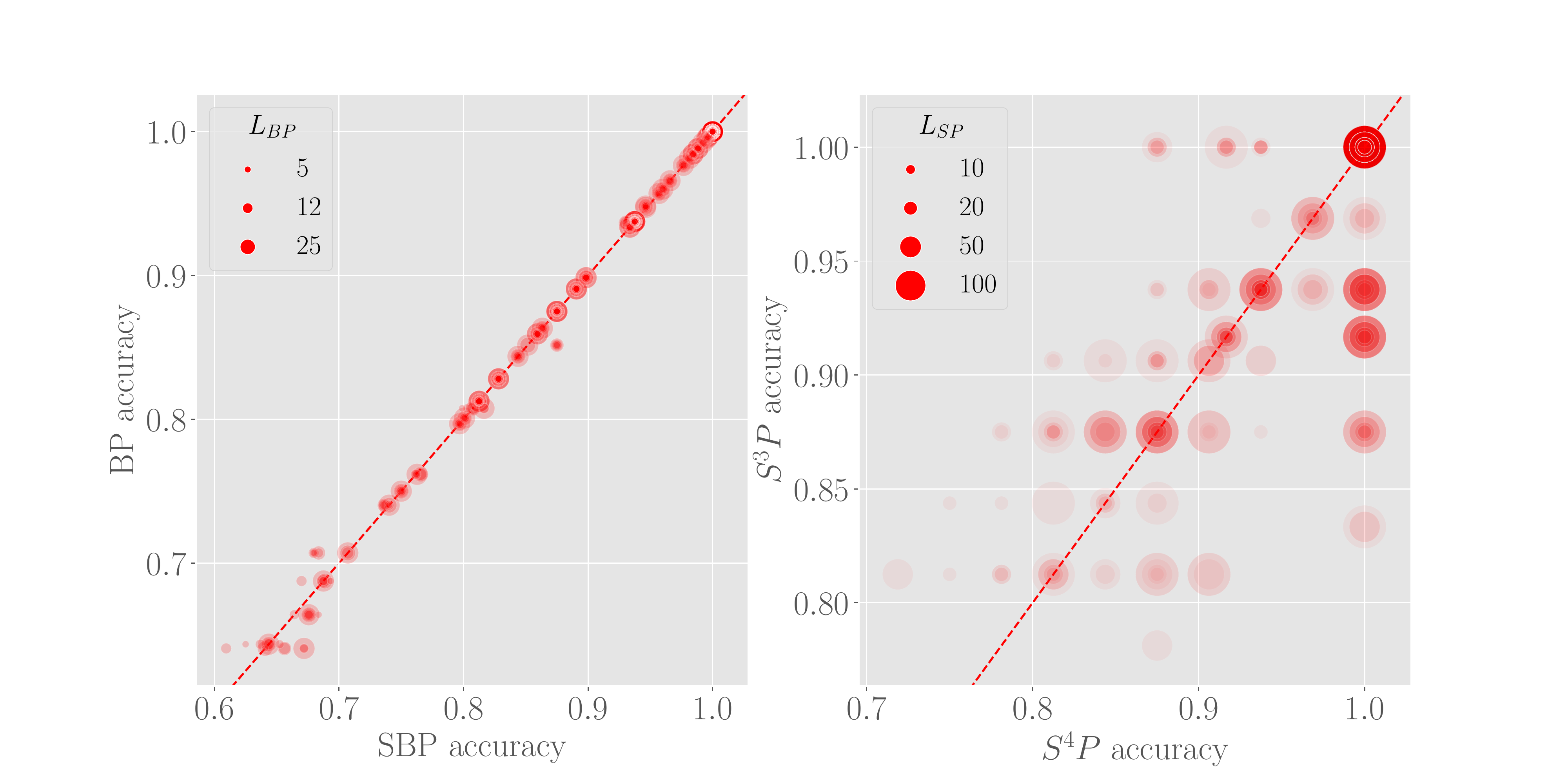}
    \caption{(left) Comparing the performance of stochastic and deterministic BP.  The red dashed line denotes a line with unit slope, and the size of each marker denotes the number of samples ($L_{BP}$) used in the MC estimate for BP.  We find the algorithms are comparable even with very small $L_{BP}$. (right) Performance of S$^4$P against S$^3$P: for larger values of samples S$^4$P is able to reach higher accuracies, due to accumulative binning errors that could arise in S$^3$P (see Appendix.~\ref{sec:exp:consistency} for details).}
    \label{fig:stoch:bp}
\end{figure}

 We first demonstrate in Fig.~\ref{fig:stoch:bp} that SBP and S$^4$P yield comparable results to the BP and  S$^3$P, respectively on a whole range of different conditions (see Appendix.~\ref{sec:exp:consistency} for details). We do not compare with exact SP since the factor to variable messages are intractable even for small problems described below. Our experiments are implemented in pytorch~\citep{paszke2017automatic}.  We use broadcasting operations on the GPU to parallelize all operations.
 
\subsection{Nonlocal solver under different regimes}
\label{sec:exp:toy_datasets}
Here,
we investigate the utility of message passing algorithms for classification tasks under carefully designed conditions. These conditions are chosen so that the difference between local and nonlocal solvers can be highlighted.
Consider what we call in our experiments the \emph{glass} dataset where $\underline{x}$ and $\underline{y}$ comprise of independent fair Bernoulli units.
We are interested in neural nets under over-parametrized and under-parametrized conditions, and the region between them. The parameter $\alpha \triangleq \frac{M}{N}$, \emph{i.e.}, the ratio of the number of training examples to the number of variables is a measure of how over- or under-parameterized the system is~\citep{mezard2009information} for this dataset. When the system is over-parameterized (small values of $\alpha$), many solutions exist. For an under-parameterized system at large values of $\alpha$, it is impossible to satisfy all the constraints. The phase transition between the two is of great interest, since the solutions are difficult to find in this region; a solver that undergoes phase transition at a higher value of $\alpha$ is therefore more powerful~\citep{Chieu2009RelaxedSP}. It is known that neural networks with full precision weights with some mild conditions on the number of hidden units and activations are easily able to learn any random assignment of labels to inputs~\citep{Zhang2017UnderstandingDL}. Therefore, in order to clearly see the over- and under-parameterization, we use a single one-layer MLP with a $\texttt{sign}$ function output. Fig.~\ref{fig:phase:linear:random}(left) depicts the performance of different algorithms on the glass dataset. Note that SNMP's accuracy drops at higher values of $\alpha$ compared with the local methods SBP and GD. The numbers on the vertical axis are the average accuracies across 20 lines. See Appendix~\ref{app:det:toy} for details.

As another benchmark, and in contrast to the glass dataset with no structure in its elements, we study the digits dataset~\citep{digitsdataset} comprising 250 images with dimensionality of 64. The results are portrayed in Fig.~\ref{fig:phase:linear:random}(right). Note that due to simplicity and structure of this dataset, a linear model is not under-parameterized and
the difference between the global and local methods are less pronounced.

As an intermediary between these two datasets, we designed the \emph{stained glass} dataset, where the inputs are sampled from fair Bernoulli distributions and the output is generated by passing the input through an MLP (with 1 or 2 hidden layers and $\texttt{sign}$ non-linearity) with random binary weights. The weights are random binary variables, and the labels, $\underline{y}$, have a functional relationship with $\underline{x}$; a relationship that is not necessarily captured by a single-layered MLP. Fig.~\ref{fig:phase:linear:random}(middle) compares the performance of different models under this dataset. Interestingly, the gap between the local and nonlocal methods is in between the two extreme cases, suggesting that nonlocal binary solvers can be of more utility in underparameterized settings.

Amongst possible approaches to using (S)GD for optimization of BNNs, we have adopted the method of \citet{bengio2016bnn}. This approach does not rely on the central limit theorem (CLT) as do methods by~\citet{peters2018probabilistic, Shayer2018LearningDW}. CLT breaks when the number of hidden units is not large enough which is the case in a one-layer network. Moreover, \citet{bengio2016bnn} use the full precision of weights in back propagation, and only use binary weights during the forward pass resulting in competitive performance.

\begin{figure}[!htbp]
    \centering
    \includegraphics[width=\textwidth]{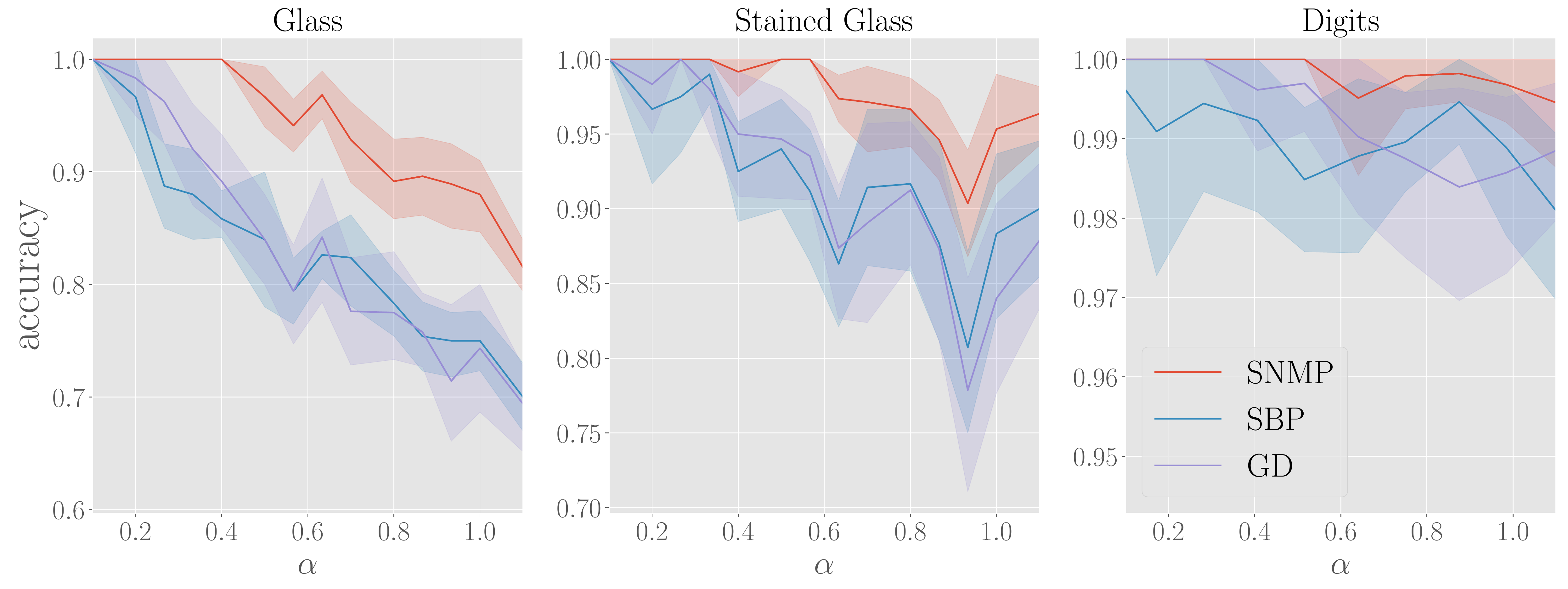}
    \caption{Comparing the performance of the proposed models and GD on several datasets under different settings (see text for detail). }
    \label{fig:phase:linear:random}
\end{figure}

\subsubsection{Benefiting from non-locality}
\label{sec:nonloc ben}

\begin{figure}[!htbp]
    \centering
    \includegraphics[width=\textwidth]{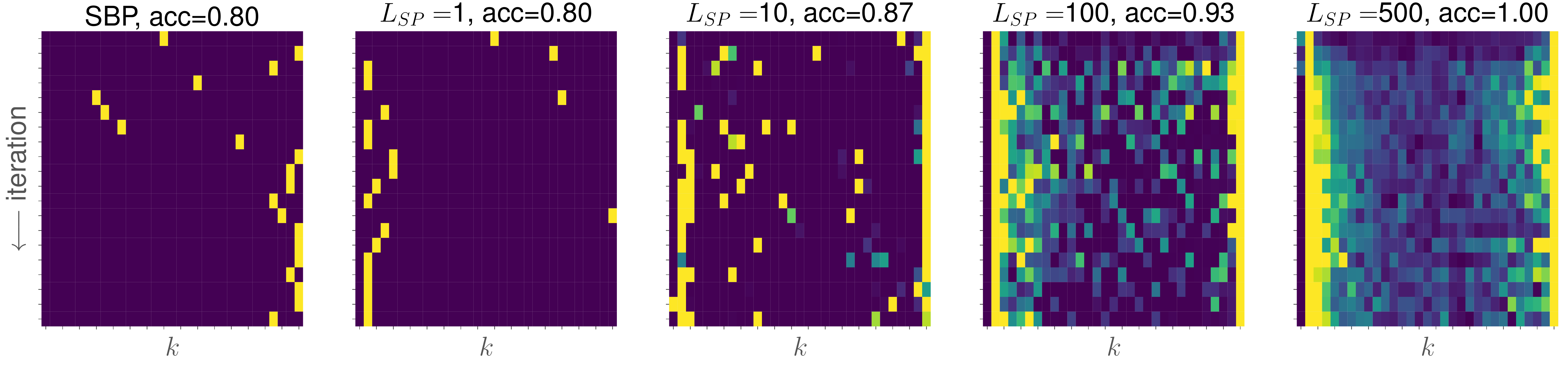}
    \caption{Visualizing the evolution of the stochastic SP marginals, for different sample sizes ($L_{sp}$).  BP is presented for contrast.  Each box represents the time evolution of the SP marginal for a given variable.  While BP can only track (and converge to) one candidate solution, SP is free to explore and propagate multiple candidates non-locally.  As $L_{sp}$ increases S$^4$P can find better configurations which are inaccessible to BP.  Best viewed in colour.}
    \label{fig:sp_bin_hopping}
\end{figure}

One of the specific advantages of using SP over a local solver like BP or GD is its ability to efficiently consider a large number of candidate solutions.  Because SP is able to track and aggregate information from potentially an exponential number of BP fixed points~\citep{Ravanbakhsh2014RevisitingAA}, it can often find solutions that are inaccessible to a single SBP or GD run.  We demonstrate this in Figure \ref{fig:sp_bin_hopping}, which shows the evolution of a given $S^4P$ marginal over iterations of the algorithm.  Each box represents a model, initialized in the same way and trained on the same dataset, but by varying the number of samples used in stochastic SP, we can control the ability of the algorithm to explore the solution space.  The extreme case, $L_{sp}=1$, is also presented, and contrasted with SBP.  In the case of BP, we see the beliefs hop around until they eventually converge on a single solution which achieves sub-optimal accuracy.  As we increase the sample size, more and more mixing is possible, which translates to more accurate configurations that were previously inaccessible to the model.

\subsection{MNIST dataset}
\label{sec:exp:mnist}
\begin{figure}[!htbp]
    \centering
    \includegraphics[width=\textwidth]{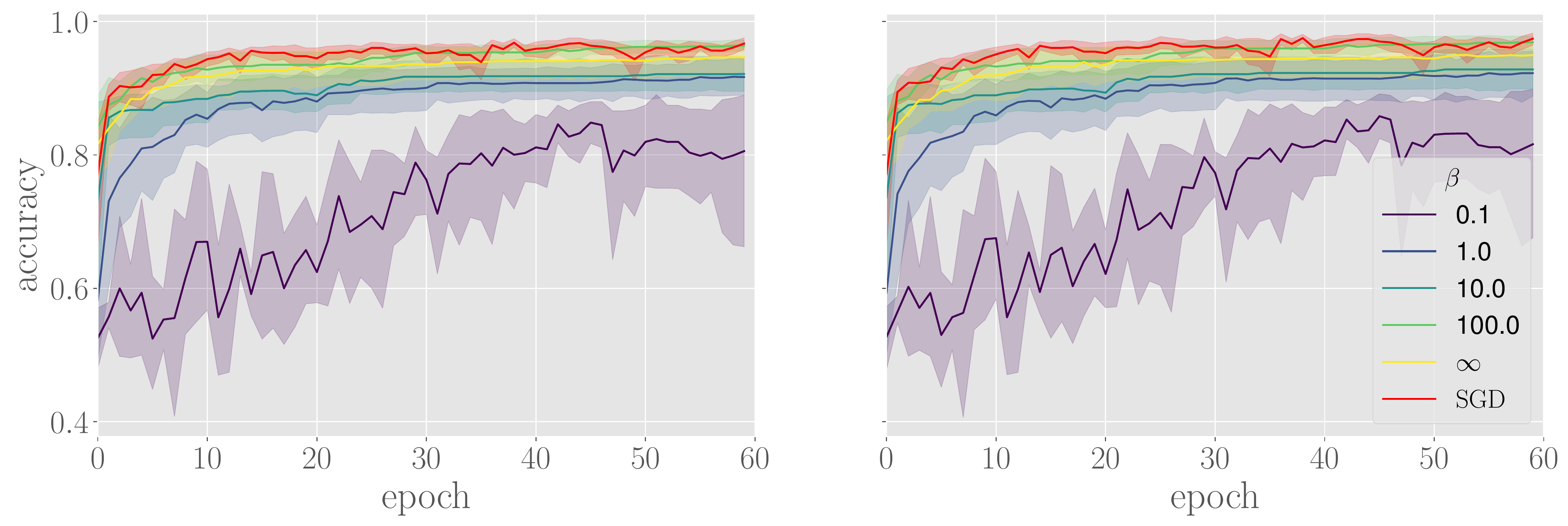}
    \caption{Training curves on full MNIST dataset for multi-layer convolutional network trained with Max-Sat, using a mini-batch approach.  We observe that higher $\beta$ leads to a higher accuracy score, though noticeably $\beta_{\infty}$ is not the optimal value.  Shown for comparison is SGD.  The figure on the left is the training accuracy and on the right is the test accuracy.}
    \label{fig:maxsat:convnet}
\end{figure}

We study the performance of our SNMP against gradient-based methods
using a binary classification task on MNIST dataset. We have chosen linear, MLP and convolutional networks. See Appendix~\ref{app:det:mnist} for details.
The phase transition diagrams for both the test and training sets are plotted for the linear, MLP and convolutional networks in Fig.~\ref{fig:phase:linear:random}.  
In order to scale our vectorized implementation to a larger number of MC samples, we use a mini-batch version of message passing as delineated in Appendix.~\ref{app:det:mnist}. This allows running the algorithm on all examples of MNIST (from the given classes). Fig.~\ref{fig:maxsat:convnet} compares the results of running SNMP with different values of $\beta$ with the SGD algorithm. The best performance is achieved at higher values of $\beta$, but not at $\beta = \infty$. The results are comparable with SGD.
\begin{figure}[!htbp]
    \centering
    \includegraphics[width=\textwidth]{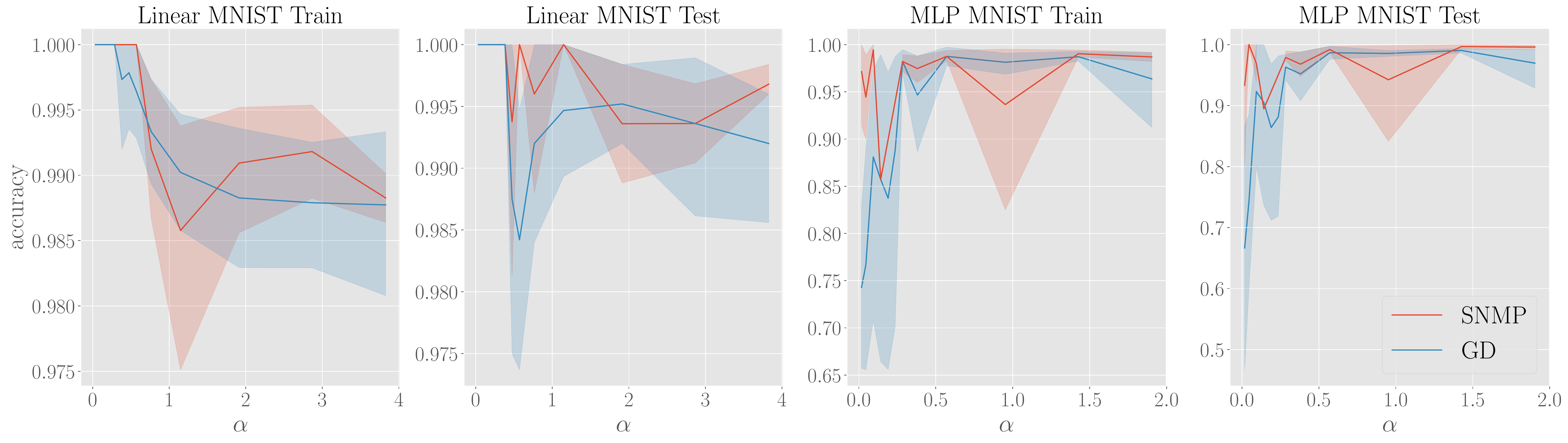}
    \caption{Comparing performance of the different solvers on the MNIST dataset.  The two leftmost plots correspond respectively to the training and test performance of a linear model, whereas the latter two are obtained from a two-layer neural network with sign-function activations.}
    \label{fig:phase:linear:random}
\end{figure}

\section{Conclusions and future work}
We extended the framework of message passing to arbitrary factors with high degrees and used this approach to obtain nonlocal solvers for binary neural networks. Our studies reveal that using powerful nonlocal solvers can be beneficial in underparameterized systems. These systems are interesting in energy-constrained environments where vast computational resources are not available. We showed that our models perform competitively with gradient based methods and are able to access parts of the solution space that is not directly accessible to local methods. Application and extension of these models to large scale datasets is a prime direction for our future studies.

\clearpage

\bibliographystyle{plainnat} 
\bibliography{references}  

\clearpage
\newpage
\title{Supplementary material}
\maketitle

\appendix
\renewcommand\thefigure{\thesection.\arabic{figure}}   
\section*{Supplementary Material for Nonlocal optimization of binary neural networks}

\section{Consistency between stochastic and exact models}
\label{sec:exp:consistency}

We gauge the accuracy of SBP and S$^4$P against BP and S$^3$P on a large number of problems and range of parameters for which BP and S$^3$P are tractable.  More specifically, we fix $M = 2^N$, and  sample all combinations from a grid formed by:

\begin{itemize}
    \item $N \in \{4,6,8,10\}$
    \item $\gamma \in \{0.2, 0.5, 0.8\}$
    \item $L_{BP} \in \{5, 8, 10, 12, 15, 20, 25, 50\}$
\end{itemize}

SBP and BP are run on the resulting problems.  A comparison between the accuracies can be seen in Fig.~\ref{fig:stoch:bp}(left), where we find nearly identical agreement even when a small number of samples are used. It can also be seen that SBP is relatively insensitive to the choice of $L_{BP}$, \emph{i.e.}, the number of MC samples.

In a similar way, to compare S$^3$P and S$^4$P, we fix $N=20$ and sample all combinations from a grid of:

\begin{itemize}
    \item $K \in \{31, 51, 101, 201\}$
    \item $\gamma \in \{0.2, 0.5, 0.8\}$
    \item $L_{BP} \in \{5, 10, 20, 50, 100\}$
    \item $L_{SP} \in \{5, 10, 20, 50, 100\}$
    \item $M \in \{8, 12, 16, 32\}$
\end{itemize}

A comparison between these accuracies is also present on the right side of Fig.~\ref{fig:stoch:bp}. Two important points are noted: first, higher numbers of MC samples lead to better S$^4$P performance. Second, quite interestingly, S$^4$P usually outperforms S$^3$P, due to the binning errors that accumulate if all the the exponential combinations of bins are considered.

\section{Stochastic factor to variable update in SP}
\label{sec:app:var2f_sp}
Similar to the derivations in the main text, the variable to factor message can be written as:
\begin{equation}
    \tilde{S}_{i \rightarrow I}({p}_{i \rightarrow I}= p_k) =
    \mathbb{E}_{S({p}_{\partial i\setminus I})}\Bigg[
       \mathbb{I}\left[ p_k = {p}_{i\rightarrow I}(.; p_{\partial i\setminus I \rightarrow i})\right]
   {p}_{i\rightarrow I}(.; p_{\partial i\setminus I \rightarrow i})
   (\emptyset)
    \Bigg],
    \label{eq:SP_f2var_stoch}
\end{equation}
where ${p}_{k} \in [0, 1]$ is one of the possible $K$ values that ${p}_{i \rightarrow I}$ can take. ${p}_{I\rightarrow i}(.; p_{\partial I\setminus i \rightarrow I})
   (\emptyset)$ is calculated directly using Eq.~\ref{eq:var2f_orig}. Algorithm~\ref{alg:sp_var2f} shows how this approach works. The algorithm receives variable to Factor surveys (SP messages) as input $\{{S}_{i \rightarrow I}({p}_{{i \rightarrow I}})  \}_{\{i \in \partial I\}}$.
 The MC estimate is done over $L_{sp}$ samples (lines 1-9). A  variable to factor probability distribution is sampled from the surveys incident on variable $i$. This probability distribution is fed into the   \texttt{BPVar2F} subroutine (Eq.~\ref{eq:var2f_orig}) to obtain the normalized and unnomarlized probabilities (line 3). These unnormalized probabilities are used to calculate the local partition function for each variable (line 5), whereas the normalized probability is used to find the bin $k$ that corresponds to the given probability (line 6), where $\texttt{bin}(p_k) \triangleq k$. The message corresponding to this bin is subsequently updated (line 7). Finally, the normalized messages are calculated (lines 10-14).

\begin{algorithm}[!htbp]
  \SetAlgoLined
\KwIn{Normalized messages $\{{S}_{I \rightarrow i}({p}_{{I \rightarrow i}})  \}_{\{I \in \partial i\}}$ incident on variable $i$}
\KwOut{Normalized outgoing messages $\{{S}_{i \rightarrow I}({p}_{{i \rightarrow I}})  \}_{\{i \in \partial I\}}$ from variable $i$ }
\For{$m \in \{1, \dots L_{\text{sp}}\}$}
{
    ${ p}_{ \partial i \rightarrow i} \sim 
    \prod_{J \in \partial i}{S}_{J \rightarrow i}({p}_{J \rightarrow i})
    $

    ${ p}_{i \rightarrow I}(w_{ i}=1),  {\tilde p}_{i \rightarrow I}(w_{i}=1),  {\tilde p}_{i \rightarrow I}(w_{i}=0)$ $\leftarrow$
    \texttt{BPVar2F}$({ p}_{ \partial i \rightarrow i})$
    
        \For{$I \in \partial i$}{
        
        ${p}_{i \rightarrow I}(\emptyset) \leftarrow {\tilde p}_{i \rightarrow I}(w_i=0)+{\tilde p}_{i \rightarrow I}(w_i=1)$
        
        $
        {k} \leftarrow \texttt{bin}({ p}_{i \rightarrow I}(w_i=1))
        $
        
        $
        \tilde{S}_{i \rightarrow I}(p_k) \leftarrow \tilde{S}_{i \rightarrow I}(p_k) + {p}_{i \rightarrow I}(\emptyset)
        $

 }
 }
 \For{$I\in \partial i$}{
 \For{$k \in \{1, ..., K\}$}{
 ${S}_{i \rightarrow I}({p}_{k}) \leftarrow \frac{\tilde{S}_{i \rightarrow I}({p}_{k})}{\sum_{k=1}^{K} \tilde{S}_{i \rightarrow I}({p}_{k})}$
 }
 }
 \caption{SP Stochastic variable to factor subroutine. 
 }
 \label{alg:sp_var2f}
\end{algorithm}

\section{Experimental and implementation details}
\label{sec:app:hps}
\setcounter{figure}{0} 

\subsection{General considerations when tuning hyperparameters}
\label{app:det:general}

Here, we summarize the different hyperparameters used in our solver, and their recommended settings.

\begin{itemize}
    \item $L_{BP}$: the number of samples used for SBP. As shown on the left of Fig. \ref{fig:stoch:bp}, one can get away with typically very small number of samples.  In all our experiments, we set $L_{BP}=5$.
    \item $L_{SP}$: the number of samples used for S$^3$P and S$^4$P.  As shown in Fig. \ref{fig:sp_bin_hopping}, this parameter is crucial for proper exploration of the space, and using higher $L_{SP}$ will typically improve the final accuracy.  From our own tuning, we empirically find that one wants at least $L_{SP} > 10 L_{BP}$.    If the memory space is available, $L_{SP}$ is one of the first things to tune.  In all our experiments we have at least $L_{SP}=100$, though we go up to 500 when we can afford it (as done for Fig. \ref{fig:sp_bin_hopping}).
    \item $K$: the number of bins used for SP.  A low value results in insufficient resolution to track the different BP messages.  However, adjusting $K$ offers  little after a certain threshold.
    \item $\gamma$: the damping factor.  We use damping in our updates to make the system more stable~\citep{murphy_graph}. If the previous value of a message (variable to factor or otherwise) is given by $\mu_t$, then with damping factor $\gamma$ after computing updates $\mu$, the new message becomes:  $\mu_{t+1} =(1 - \gamma) \mu_t + \mu$.  When $L_{SP}$ is high, $\gamma$ can afford to be higher, and converges more quickly.  For SBP we typically set $\gamma = 0.2$, while for S$^4$P we use $\gamma = 0.8$.
    \item $\beta$: the inverse temperature for Max-Sat.  Higher values of $\beta$ have a positive impact on the variance encountered during training, especially when combined with tuning $\gamma$.  Fig. \ref{fig:maxsat:convnet} provides a good illustration of the impact of $\beta$ on training.
\end{itemize}

\subsection{Details for experiments in Sec.~\ref{sec:exp:toy_datasets}}
\label{app:det:toy}
The message passing algorithms converge much faster (usually less than 5 epochs) than gradient based methods. However, we run all the algorithms for a higher number of epochs (20) so that SGD can reach its solution. We fine-tune the learning rate of SGD for optimum performance. The best performance was obtained with a learning rate of $0.1$.

For both SBP and S$^4$P, we set $L_{BP}=5$.  For S$^4$P, we use $L_{SP}=100$ for all experiments, with $K=201$. In all our experiments, we keep the number of variables fixed and adjust the number of training examples.

For the glass dataset, we set $N=10$ and let $M$ span $[5...50]$ jumping in units of 5.  We use the same span of $M$ for stained glass, but set $N=31$. For the digits dataset, the linear model requires $N=64$ to match the input dimensions.  For $M$, we span the range $[5...245]$ jumping in units of 10.

For Fig.~\ref{fig:sp_bin_hopping}, we set $N=10$, $M=20$, and ran for 20 iterations with $\gamma = 0.5$.  $K$ was set to $31$ for ease of visualization.

\subsection{Details for experiments in Sec ~\ref{sec:exp:mnist}}
\label{app:det:mnist}

We test the solver's ability to handle models beyond simple linear classifiers, and test on a larger dataset, namely MNIST.  
The linear model and the MLP operate on flattened images.  In order to reduce the number of variables required for this, we use a max-pooling operation, reducing the images to size (14,14), and thus the input dimension to $14 \times 14 = 196$.  The linear model thus consists of 196 parameters, while the MLP, with a single hidden layer of size 2, has $196 \times 2 + 2 = 394$ parameters.

Finally, the conv net consists of three layers:

\begin{itemize}
    \item A Convolutional layer with 3 (3x3) filters, using a stride of 2. The activation is a simple threshold $y = (h > 0)$
    \item A second convolutional layer, with 2 (3x3) filters, also using a stride of 2, followed by threshold at 0.
    \item The outputs of the previous layer are flattened to form a vector of size (8), after which a linear layer with a sign function activation is used.
\end{itemize}

Taken together, the convolutional network has 89 parameters. Note that none of our models contain biases.

We use $\gamma = 0.5$ for the linear model and the MLP.  For the convolutional network, we adapt $\gamma$ according to the present accuracy of the model.  At the beginning of training, when accuracy is low, $\gamma$ should be high to explore the space quickly.  As the model becomes more accurate, however, $\gamma$ can be reduced to limit the variance.  As such, every 20 iterations we compute the accuracy, and set $\gamma = (1 - \textrm{accuracy})$.

As Sec.\ref{app:det:vectorized} describes, our implementation vectorizes the message computation, and performs all MC estimations in one pass.  The tradeoff is that this requires us to multiply the size of those messages in memory by $L_{SP}$.  Thus, rather than using a smaller sample size for our MC estimation, we draw inspiration from minibatch optimization and consider ways to update the messages using a smaller subset of the factors.  At a given iteration, we select a subset of the factors, and only update the factor to variable messages from those factors.  We then immediately update all of the variable to factor messages, instead of looping over the remaining factors.  

Fig. \ref{fig:maxsat:convnet} shows the training history of a three layer convolutional neural network, using the Max-Sat strategy described in Sec. \ref{sec:method:problem_setup}.  We explore $\beta$ in the range $\{0.1, 1, 10, 100, \infty\}$, where $\infty$ corresponds to regular SBP.  

\subsection{General considerations for vectorized implementations}
\label{app:det:vectorized}

For BP and SBP, the variable to factor and factor to variable messages can be represented as matrices with shape $[M,N]$, and similarly, for S$^3$P and S$^4$P the messages have shape $[M,N,K]$.  The main computational demands of the algorithm are in the generation, and subsequent evaluation, of candidate configurations for the variables.  

For generating configurations, the main use of memory is in sampling large multi-dimensional matrices.  For instance, to compute the message updates in $SBP$ described in Alg. \ref{alg:bp_f2var}, each factor requires sampling a matrix with shape $[L_{BP}, 2N, N]$, which grows quadratically with the number of variables.  We can generate the full set of samples in one pass by uniformly sampling a 4D matrix of size $[M, L_{BP}, 2N, N]$, and then performing broadcasted element-wise comparison with the variable-to-factor messages along the second and third dimensions of the matrix.  

For the factor-to-variable messages in Alg. \ref{alg:sp_f2var}, before we can sample the candidate configurations, we must first generate samples from the categorical distribution over the $K$ bins described by the variable to factor messages.  We may generate these by sampling uniform matrices with shape $[L_{SP}, M, N]$, and performing broadcasted comparisons with the cumulative distribution over the $K$ bins.  Having obtained these sample bin indices, we can look up the corresponding probabilities, which we can use to generate the sample configurations.  This proceeds in the same way as SBP, except we now sample 5D matrices of shape $[M, L_{SP}, L_{BP}, 2N, N]$.

The configurations, as sampled, consist of flattened and concatenated values for the parameters of the BNN.  We load and reshape contiguous subsets of these parameters according to the architecture of the model.  For simple linear layers, matrix mulplication can be performed by broadcasting element wise multiplication, followed by summation.  The evaluation of convolutional layers can be performed efficiently using grouped convolution.  If the convolutional layer requires filters of shape $[C_{out}, C_{in}, K_W, K_H]$, the corresponding subset of the configuration matrix can be reshaped to $[M, L_{SP}, L_{BP}, 2N, C_{out}, C_{in}, K_W, K_H]$.  To leverage the efficient grouped convolutions present in pytorch, we reshape these configurations into a matrix of shape $[B \times C_{out}, C_{in}, K_W, K_H]$, where $B = M \times L_{SP} \times L_{BP} \times 2N$.  The inputs to the layer are  reshaped to $[1, B \times C_{in}, W, H]$.  We then perform convolution using groups of size $B$ (and subsequently unpack the reshaped outputs).

With the batch version, we seek a tradeoff between memory requirements and the amount of computation.  We sample a subset of the factors of size $b$, and use the same procedure above to compute the messages to all variables. The matrices sampled in this way posses the same dimensionality.

\end{document}